# Using the Gene Ontology Hierarchy when Predicting Gene Function

**Sara Mostafavi**
Department of Computer Science,
University of Toronto

**Quaid Morris**
Banting and Best Department of Medical Research and
Departments of Computer Science and Molecular Genetics,
University of Toronto

## Abstract

The problem of multilabel classification when the labels are related through a hierarchical categorization scheme occurs in many application domains such as computational biology. For example, this problem arises naturally when trying to automatically assign gene function using a controlled vocabularies like Gene Ontology. However, most existing approaches for predicting gene functions solve independent classification problems to predict genes that are involved in a given function category, independently of the rest. Here, we propose two simple methods for incorporating information about the hierarchical nature of the categorization scheme. In the first method, we use information about a gene's previous annotation to set an initial prior on its label. In a second approach, we extend a graph-based semi-supervised learning algorithm for predicting gene function in a hierarchy. We show that we can efficiently solve this problem by solving a linear system of equations. We compare these approaches with a previous label reconciliation-based approach. Results show that using the hierarchy information directly, compared to using reconciliation methods, improves gene function prediction.

## 1 Introduction

We are interested in the problem of multilabel classification when the labels are related through a hierarchical categorization scheme and the input data is represented as a similarity metric between objects. This problem arises naturally when trying to automatically assign gene function, using controlled vocabularies like Gene Ontology (GO) (Ashburner *et al.*, 2000), Enzyme Commission (EC) number (Bairoch, 2000), and Structural Classification of Protein (SCOP) categories (Murzin *et al.*, 1995). The hierarchical multilabel classification problem, as we have posed it, arises in other application domains including automatic image segmentation and automatic web-page annotation.

In the domain of gene and protein function prediction, the input data is most naturally represented using a network whose nodes represent genes (or proteins) and whose edges are positively weighted according to the biological evidence for shared function of the connected genes. Such networks, which are called *functional association* or *function linkage networks*, can be derived from a variety of genomics and proteomics data sources, including gene expression and genetic interaction data, by using an appropriate similarity metric. For example, Pearson correlation coefficient is often used to measure similarity between gene expression profiles.

Although genes have multiple functions, existing approaches for predicting gene function typically solve a binary classification problem to identify positive genes for each function category independently (Pavlidis *et al.*, 2002; von Mering *et al.*, 2003; Lanckriet *et al.*, 2004; Tsuda *et al.*, 2005; Myers *et al.*, 2005; Mostafavi *et al.*, 2008). However, hierarchical gene classification schemes, such as GO, organize gene function categories as a directed acyclic graph (DAG) in which categories describing broader functions (e.g. eye development) are ancestors of those describing more specific functions (e.g. eye photoreceptor cell differentiation). Annotations, i.e., assignments of genes to a given category, satisfy the "true path" rule: genes annotated to a given category are also assigned to all of its ancestors; for example, an annotation in the category photoreceptor cell differentiation implies annotation in eye development.

When making an annotation, curators place genes in the most specific category supported by the available data. Often genes are annotated in internal nodes of the DAG because there is insufficient evidence to annotate genes in the most specific, i.e., leaf, categories. For example, a mouse gene can be annotated as being involved in development if mice with defective copies of that gene die as embryos. Further investigations may determine whether the gene functions in eye, heart, or brain development, warranting a more specific annotation. These internal node annotations can provide helpful hints when classifying genes in descendent categories, so long as the classification algorithm incorporates prior knowledge about the hierarchy.

Here, we introduce two new classification methods that leverage DAG-based categorization hierarchies. Both of our algorithms extend the Gaussian random fields (GRF) algorithm (Zhou et al., 2004; Zhu et al., 2003). Our interest in this algorithm stems from its success at predicting gene function compared with other binary and hierarchy-based classification schemes (Pena-Castillo et al., 2008; Mostafavi et al., 2008; Tsuda et al., 2005). Intuitively, the GRF algorithm takes as input a similarity network (here a functional linkage network) and a set of real-valued label biases and assigns real-valued discriminant scores to each node; these scores are assigned so that the linked nodes have similar discriminant scores and the discriminant score of each node is not too different from its initial label bias. Our first method, which we call *Hierarchical label propagation (HLProp)*, replicates the similarity network for each category and then links the nodes representing the same gene in parent and child categories, thus ensuring that the discriminant scores of a gene in related function categories also remain close. By applying the GRF algorithm to this new (much larger though sparsely-connected) network, we can perform multilabel classification efficiently by solving a linear system of equations. We also describe a second method, *Hierarchical label bias (HLBias)*, that uses the GO hierarchy to set label biases of genes with annotations in internal category nodes. This second approach builds on the previous work of (Eisner et al., 2005) which used the structure of the GO hierarchy to define positive and negative examples for a given category of interest.

In Section 3, we briefly review the GeneMANIA version of the GRF algorithm (Mostafavi et al., 2008); show the relationship between GRF label propagation and Gaussian inference; and then use that relationship to justify our two extensions: HLBias and HLProp. In Sections 4 and 5, we compare these two methods, and two simplifications of HLProp, a reconciliation method called Isotonic Regression (Obozinski et al., 2008), and unaugmented binary classification using GeneMANIA GRF (Mostafavi et al., 2008). We evaluate performance in two settings, test and novel settings, using the data from the MouseFunc challenge (Pena-Castillo et al., 2008). The test setting evaluates each classifier in a cross-validation framework in which the GO annotations of test genes are completely hidden. The novel setting evaluates the performance in the task of predicting new annotations given the state of the GO annotation database from the previous year. Many annotations in the updated GO database are refinements of pre-existing internal node annotations.

## 2 Previous Work

To date, most classification algorithms that make use of the GO hierarchy have built on top of binary classification schemes; standard structured output classification algorithms (Taskar et al., 2003) are difficult to employ here due to the size the classification problem and the non-trivial tree-width of the GO hierarchy. Augmented binary classification schemes include cascaded classification (Kiritchenko et al., 2004) which trains one binary classifier for each category to predict whether annotation in the child category is warranted given annotation in the parent category (or categories). Annotation predictions are made by querying classifiers in a cascade from the root down. An alternative approach is to independently train binary classifiers for each category and then to reconcile their predictions so that the true path rule is enforced. For example, (Barutcuoglu et al., 2008) reconciled predictions of binary SVMs for 100 GO categories (a subset of about 2,000 well-annotated GO categories) by using a Bayesian network to model the GO hierarchy over the considered categories, (Obozinski et al., 2008) extended this approach to the entire GO hierarchy using approximate inference.

Obozinski and colleagues (Obozinski et al., 2008) compared ten reconciliation methods including reconciliation with a Bayesian networks (as done in (Barutcuoglu et al., 2008)), Isotonic Regression, and a cascade of classifiers approach (Cascade Logistics Regression) on the MouseFunc benchmark. Surprisingly, many reconciliation methods did not significantly improve prediction accuracy, however, the performance of Isotonic Regression was better than that of Cascade Logistic Regression and other reconciliation methods. As such, in this report, we restrict our comparisons to Isotonic Regression. Other hierarchy-aware gene function prediction schemes that have only been applied to tree-structured hierarchies, e.g. (Shahbaba & Neal, 2006), will not be considered here.

## 3 Algorithm

We assume that we are given as input a weighted, undirected network represented as a positive, symmetric affinity matrix $W = W^\mathsf{T}$, $w_{ij} \geq 0$, where $w_{ij}$ indicates the strength of the evidence of co-functionality between genes $i$ and $j$; a classification hierarchy over $d$ categories denoted by $H_{d \times d}$, where $h_{mc} = 1$ indicates that category $c$ is a child of category of $m$; and a label bias matrix $Y = [\vec{y_1}, ..., \vec{y_d}]$, where $\vec{y_c} \in \{+1, k, -1\}^{n \times 1}$ consists of the label biases of all genes in the category $c$. As we will show later, the parameter $k$ is set to a scalar between -1 and +1 and reflects our prior bias on the mean of the labels of unlabeled genes. We consider a transductive setting, i.e., we assume that all input vectors are available during training; this is a natural assumption since the known gene complement of genomes is relatively stable.

Below, we first describe how to predict gene function using the GRF algorithm; we then show how to use the GO hierarchy to derive more informative label biases. Next, we show how to extend GRF when a classification hierarchy is available. Finally, we describe other approaches for making gene function predictions with a classification hierarchy.

### 3.1 Predicting a single gene function

To predict gene function, we use the Gaussian random fields algorithm (Zhu et al., 2003; Zhou et al., 2004) to assign a discriminant score $f_i \in [-1, 1]$ to each node (protein) $i$ in the network. These discriminant scores can then be thresholded to classify the genes. Below, we write the Gaussian random fields algorithm in the following general form:

$$
\begin{aligned}
\vec{f}^* &= \arg\min_{\vec{f}} \sum_{i=1}^{n} \sigma_i (y_i - f_i)^2 + \sum_{i,j=1}^{n} w_{ij}(f_i - f_j)^2 \\
&= \arg\min_{\vec{f}} (\vec{f} - \vec{y})^\mathsf{T} \Sigma (\vec{f} - \vec{y}) + \vec{f} L \vec{f} \\
&= (\Sigma + L)^{-1} \Sigma \vec{y} \quad (1)
\end{aligned}
$$

where $\vec{\sigma} = [\sigma_1, ..., \sigma_n]^\mathsf{T}$ are model parameters, $\Sigma$ is a diagonal matrix with $\Sigma_{ii} = \sigma_i$, $L = D - W$ is the graph Laplacian and $D$ is a diagonal matrix with $D_{ii} = \sum_j w_{ij}$. The above objective ensures that the discriminant scores remain close to their initial labels (first term in (1)) and that the discriminant scores of linked genes (as indicated by $w_{ij} > 0$) are similar to each other (second term in (1)).

To ensure that $\Sigma + L$ is invertible, we can set $\sigma_i > 0$ and thus ensure that $\Sigma + L$ is diagonally dominant. However, to solve for $\vec{f}$, we only need to solve a linear system of equations $(\Sigma + L)\vec{f} = \Sigma^{-1}\vec{y}$, which we can do with various existing fast iterative solvers (Nocedal & Wright, 2006). Here, we use the conjugate gradient (CG) algorithm which is well-suited to this problem because our coefficient matrix is very sparse and the CG iterations only require us to take matrix-vector products with the coefficient matrix.

The solution to (1) can also be interpreted as the maximum a posteriori (MAP) estimate of $\vec{f}$, where the observations $\vec{y} \sim N(\vec{f}, \Sigma^{-1})$ with a prior on $\vec{f}$, namely $\vec{f} \sim N(\vec{0}, L^{-1})$ where $N(\vec{\mu}, K)$ is the normal distribution with mean $\vec{\mu}$ and covariance matrix $K$ (Shental et al., 2008).

This relationship suggests that the label bias $y_i$ can be viewed as a noisy estimate of a soft label $f_i$ with the regularization parameter $\sigma_i$ as the precision of the estimate and the weights $w_{ij}$ as inverse prior covariance between $f_i$ and $f_j$. This interpretation suggests that a node's label bias should reflect our prior beliefs about its label. We have previously shown (Mostafavi et al., 2008) in unbalanced problems, we can achieve a large gain in classification performance by setting the label bias $k$ of unlabeled nodes to be the $k = \frac{n^+ - n^-}{n^+ + n^-}$, the mean of the labels of the labelled nodes. In the following section, we describe how we use the GO hierarchy to set the label bias for nodes which we have previously labelled as negative.

### 3.2 Hierarchical label bias

In Hierarchical label bias (HLBias), we use the GO hierarchy directly to set the initial label bias of non-positive genes. HLBias builds on the previous work of (King et al., 2003) and (Eisner et al., 2005). In particular, King and colleagues (King et al., 2003) used a gene's annotations as a feature vector for predicting additional annotations for the given gene. Eisner and colleagues (Eisner et al., 2005), used the structure of the GO hierarchy to define appropriate negative examples for predicting a given gene function: they used as negatives all genes with initial annotation (i.e., before calculating the transitive closure using the true path rule) in neither descendant nor ancestral categories.

In our approach, we use a gene's previous annotations to estimate our prior bias that it will be annotated to a given category of interest. To do so, when predicting category $c$, we first use as negatives all genes that are annotated in any sibling category of $c$. We assign this negative label because genes are rarely annotated in more than one child of the same parent category. For other genes $i$ with an annotation in an ancestral category $a$ of $c$, we set $y_{ic} = 2 \times \frac{n^+_{ac}}{n^+_a} - 1$ where $n^+_a$ is the number of positive examples in category $a$ and $n_{ac}^+$ is the number of positive examples in category $a$ that were also annotated in category $c$; this initial label bias is proportional to the probability of a gene being annotated to category $c$ given its annotation in category $a$. For a gene $i$ with multiple annotations, we set $y_{ic}$ to its mean value. Having set these label biases, we then solve for discriminant values independently.

### 3.3 Hierarchical label propagation

Given a matrix of label biases $Y_{n \times d}$ and a hierarchical classification scheme represented by $H_{d \times d}$, we solve for discriminant values for all $d$ classes simultaneously. To do so, we solve the following problem:

$$F^* = \arg\min_{f_{ic}} \sum_i^n \sum_c^d \sigma_i(y_{ic} - f_{ic})^2 + \quad (2)$$

$$\sum_c^d \sum_{i,j}^n w_{ij}(f_{ic} - f_{jc})^2 + \lambda \sum_i^n \sum_{c,m}^d h_{mc}(f_{im} - f_{ic})^2$$

where $F = [\vec{f}_1, ..., \vec{f}_d]$ and $h_{mc}$ denotes the relationship between category $m$ and $c$, and $\lambda, \sigma_i$'s are the regularization constants. Without the third term (by setting $\lambda = 0$), equation (2) corresponds to solving $d$ independent binary classification problems. The third term encourages the discriminant values of a gene in two related function categories to be similar to each other (see Figure 1 for an example).

In this work, we use the GO hierarchy to define $H$. GO is a DAG (directed acyclic graph), however, as we will discuss later, to ensure that problem (2) is convex, we treat GO as an undirected graph. In particular, $h_{cm} \in \{0, +1\}$ represent the parent child relationships in GO: we set $h_{cm} = h_{mc} = 1$ if $m$ is a parent of $c$ in the GO hierarchy and 0 otherwise. In addition, in our experiments we set $\lambda$ and each $\sigma_i$ to a fixed value of 1 and so we drop these constants from our subsequent equations.

#### 3.3.1 Optimization

We can solve for $F^*$ by solving the following problem:

$$\begin{aligned} F^* &= \arg\min_F \text{ trace}(F^\mathsf{T} F - 2F^\mathsf{T} Y) \\ &+ \text{ trace}(F^\mathsf{T} L F) + \text{trace}(FGF^\mathsf{T}) \quad (3) \end{aligned}$$

where $G$ is the graph Laplacian of $H$, $G = V - H$, $V = \text{diag}(v_{cc})$, and $v_{cc} = \sum_m^d h_{mc}$.

Differentiating equation (3) with respect to $F$, we get the matrix equation $(I + L)F + FG = Y$. Equivalently, we can find $F$ by solving a large sparse linear system: $A(\text{vec}(F)) = \text{vec}(Y)$, where $\text{vec}(Y)$ is an operator that stacks the columns of $Y$ atop of each other,

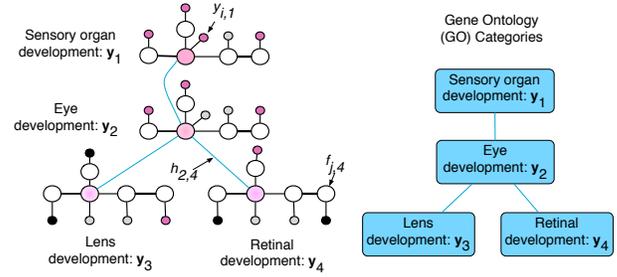

Figure 1: A graphical example of our model. On the left figure, there are four identical networks over four genes (nodes); the association between different genes is shown in black edges. The color of the smaller nodes attached to each gene represents the initial label of the gene. If we wish to predict which genes are involved in *eye development*, we need to consider other related categories (as shown on the right). In our modification, we introduce an edge between the same gene (blue edges) in the different networks (in this figure, we have only shown blue edges for one gene); these edges will encourage the discriminant value of the same gene (depicted as the color of the bigger nodes) in related categories to be similar.

$A_{(n \times d) \times (n \times d)} = (I_{d \times d} \otimes (I + L) + G \otimes I_{n \times n})$, and $\otimes$ denotes the Kronecker matrix product. As an example, the matrix $A$ that corresponds to the example in Figure 1 can be represented as:

$$A = \begin{pmatrix} (2I+L) & -I & 0 & 0 \\ -I & (3I+L) & -I & -I \\ 0 & -I & (2I+L) & 0 \\ 0 & -I & 0 & (2I+L) \end{pmatrix}$$

In general, A can be represented as a block matrix with diagonal blocks $A_{ii} = (I + L + v_{ii}I)$ and non-diagonal blocks $-h_{ij}I$.

When $H$ is symmetric, then $A$ is also symmetric. Furthermore, since A is diagonally dominant with positive diagonals, A is symmetric positive definite (SPD) and thus invertible. However, with large $d$ (number of gene function categories) and $n$ (number of genes), constructing A may be infeasible. Instead, we can solve for the $\vec{f}_c$'s iteratively: given all $\vec{f}_c$'s for $c \neq m$, we can solve for $\vec{f}_m$ by solving the system of linear equations: $(I + L + v_{mm}I)\vec{f}_m = \sum_c^d h_{cm}\vec{f}_m + \vec{y}_m$. In our setting, problem (3) is convex and we can calculate $F^*$ by iteratively updating $\vec{f}_m$'s; we have empirically observed that we need 10 or fewer iterations to solve each $\vec{f}_m^*$ when there are approximately 50 GO categories that are related to each other.

### 3.4 Other approaches for using the GO hierarchy

Here, we compare gene function prediction using HLBias, HLProp, two heuristics approaches based on HLProp (Down- and Up- propagation), and Isotonic regression, a method for reconciling the predictions of independent classifiers to satisfy the true-path rule.

*Down- or Up-propagation.* Both of our heuristics are inspired by the observation that if the classification hierarchy were a polytree, then we would be able to solve for F* using a belief propagation-like message passing algorithm which passes a single message in each direction along parent and child category link and these messages are used to calculate the label biases for the message recipient. In Up-progation, these messages are only passed from child categories to parent categories and in Down-propagation, messages are only passed from parents to their children. In particular, in Down-propagation, we first calculate $\vec{f}_r^*$, the vector of discriminant scores for genes in the root category using equation (1), then the root passes $\vec{f}_r^*$ as a message to each of its child categories $c$ who set their label biases to be $\vec{y}_c + \vec{f}_r^*$, where $\vec{y}_c$ is the vector of initial label biases for category $c$. We then apply this procedure recursively from parent to child, summing together all parent messages before calculating label biases whenever a node has multiple parents. Up-propagation is similar though messages are passed from child nodes to parent nodes.

*Isotonic Regression (IR).* Given a set of independent predictions for a gene $i$ in $d$ categories $\vec{x}_i = [x_1, ..., x_d], x_c \in \Re$, IR (Barlow *et al.*, 1972) solves the following problem:

$$\arg\min_{z_c} \sum_c^d (x_c - z_c)^2 q_c$$
$$\text{subject to } z_m \geq z_c \ \forall (m, c) \in H$$

where $q_c$'s are the parameters. We apply IR to the discriminant values obtained using Gaussian random fields algorithm to each function category independently. In our experiment, as done in (Obozinski *et al.*, 2008), we set $q_c = 1$ for $c = 1, ..., d$ and solve IR heuristically, using the generalized PAV algorithm (GPAV) (Burdakov *et al.*, 2006).

## 4 Methods

*Regularization parameters* In our experiments, we set all of the precisions $\sigma_i = 1, i \in \{1, ..., n\}$, i.e. $\Sigma = I$ and we set the regularization constant in equation (2) to $\lambda = 1$.

*Benchmark data* We use the mouse benchmark data of (Pena-Castillo *et al.*, 2008) which consists of ten genomics and proteomics datasets. We first represent each data source as a similarity network using Pearson correlation coefficient ($r$) and sparsify each network by setting to zero any interaction that is not among the top 50 highest $r$ values for either gene. To ensure that all interactions are positive (i.e. $w_{ij} \geq 0$), we only consider positive interactions ($r > 0$) and set to zero all the negative interactions. We normalize each network: $\tilde{W} = D^{-1/2}WD^{-1/2}$. We then combine all the networks, by simply adding them together to obtain a single functional linkage network and re-normalize this composite network.

*Gene Ontology* To evaluate gene function prediction we use GO function categories for *M. musculus*. We use the true-path rule to associate each gene with all of its functions, i.e. if a gene is annotated in a child category, we consider it to be annotated in all of its ancestor categories. In addition, we remove all annotations that were only supported with "inferred from electronic annotation" (IEA) evidence code; IEA annotations are the only annotations made based on previous computational predictions which have not been reviewed by a curator (Ashburner *et al.*, 2000).

*Evaluation* As in (Pena-Castillo *et al.*, 2008), we evaluate our predictions based on two sets of genes: (a) test genes (i.e. cross-validation) and (b) novel genes. For predicting test genes, we perform 3-fold cross-validation on 2,634 GO biological process categories which have between 3 and 300 annotations (GO association file download on September 1, 2007). Note that when constructing our hierarchy from which we derive the HLbias, only GO categories in this set are included. For predicting novel genes, we train on the 2007 GO association file and then we evaluate our performance in a "real-life" setting by comparing our predictions to the updated GO association file (GO association file download on September 1, 2008). In particular, for predicting novel genes, we restrict our evaluation to 903 GO categories that obtained three or more new annotations since 2007. We report the performance on predicting test and novel genes in terms of error as measured by 1-area under the ROC (Receiver Operating Characteristic) curve (1-AUC). The AUC under the ROC curve (Fawcett, 2006) corresponds to the probability that a random positive instance will be scored higher than a random negative instance. In addition to AUC of ROC, we investigate the performance in predicting novel genes as measured by AUC of precision recall curve.

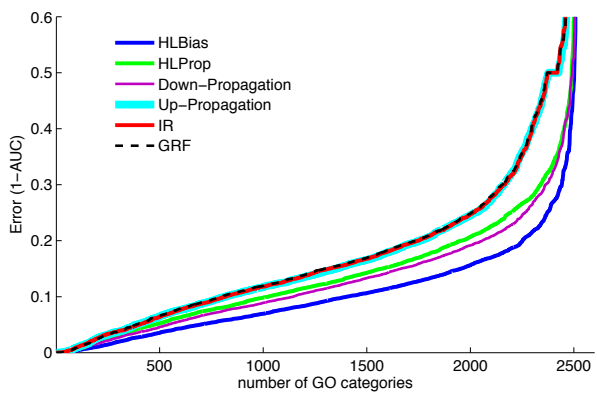

Figure 2: Cumulative performance (error) of various methods in predicting the function of test genes in 2,634 GO categories (i.e. using 3-fold cross-validation on 2008 GO annotation file).

## 5 Experimental Results

We first show the performance on test genes and then focus our analysis in predicting novel genes.

### 5.1 Predicting test genes

Figure 2 shows the cumulative distribution of the error (1-AUC of ROC) of the Hierarchical label bias, HL-Prop, Down- and Up-propagation, Isotonic Regression (IR) and Gaussian random fields (GRF) algorithms. Table 1 summarizes the mean and median error of each method.

Table 1: Mean and median error in predicting test genes in 2,634 GO categories. The last column shows the standard error in the estimate of the mean error.

| Approach | mean | median | SE |
| --- | --- | --- | --- |
| Hierarchical label bias | 0.1035 | 0.0881 | 0.0016 |
| HLProp | 0.1382 | 0.1200 | 0.0020 |
| Down-Propagation | 0.1282 | 0.1109 | 0.0019 |
| Up-Propagation | 0.1735 | 0.1405 | 0.0028 |
| IR | 0.1745 | 0.1438 | 0.0028 |
| GRF | 0.1760 | 0.1464 | 0.0028 |

HLProp, HLBias, and Down-propagation considerably reduce the error in gene function prediction. Specifically, despite being the simplest method, Hierarchical label bias achieves the best overall performance in terms of 1-AUC of ROC curve (Table 1). Note that in the cross-validation setting, for the Hierarchical label bias method, the test genes are labeled as unknowns and the initial label of other non-positive genes is set according to their previous annotations in the GO hier-archy (see Section 3.4). One explanation for the better performance of HLBias compared to HLProp is that, in using HLBias, genes that have an incomplete annotation in an ancestral category more directly influence the discriminant scores of their linked genes. This is because the initial label bias of a gene essentially needs to be *propagated* through a minimum of two edges to effect the label bias of a test gene; in HLProp the incomplete annotation information needs to be *propagated* through a minimum of three edges to effect the discriminant score of a test gene (see Figure 3 for a pictoral description). This explanation may also offer an insight into why Down-Propagation performs better than Up-Propagation.

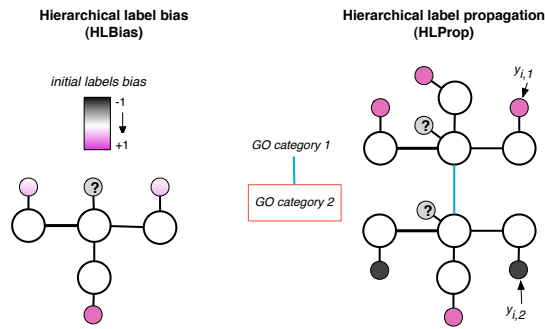

Figure 3: An example illustrating the difference between HLBias and HLProp in assigning discriminant scores to a test gene. The test gene is depicted by the node whose initial label bias is a question mark. On the right figure, the nodes in the same column depict the same gene in different categories; only one of the five blue edge representing the edges $h_{a,c}$'s is shown. In HLBias, when predicting GO category 2, the two neighbouring nodes of the test gene have a more positive label bias. In HLProp, the previous annotations needs to be propagated through more edges to effect the discriminant score of the test gene.

In addition, we observed that IR and Up-propagation do not significantly improve the performance. This result is consistent with the observations in (Obozinski et al., 2008) that most reconciliation methods often perform similar to the baseline of independent predictions.

### 5.2 Predicting novel genes

Here we report the performance in predicting novel gene function; in particular, to evaluate the performance on a given category, we use newly annotated genes as positives and all other genes (excluding previously annotated genes, that is, those annotated in the 2007 GO file) as negatives. In predicting novel genes with the HLBias method, for a given GO category, we adjust the initial label of all except the positive genes (which have an initial label bias of +1 according to the 2007 GO file) by using the incomplete annotation information in GO.

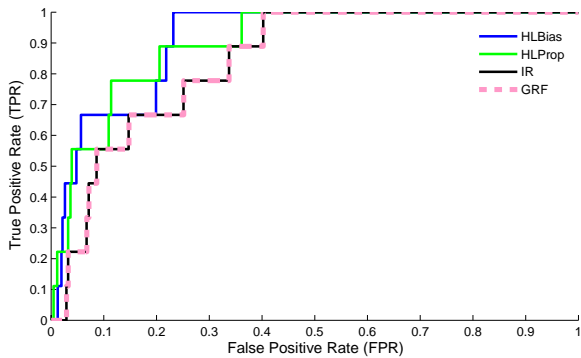

Figure 4: The ROC curve for predicting the GO category "Osteoblast Differentiation" which had 26 annotations in the 2007 GO association file and 9 new annotations in the 2008 version.

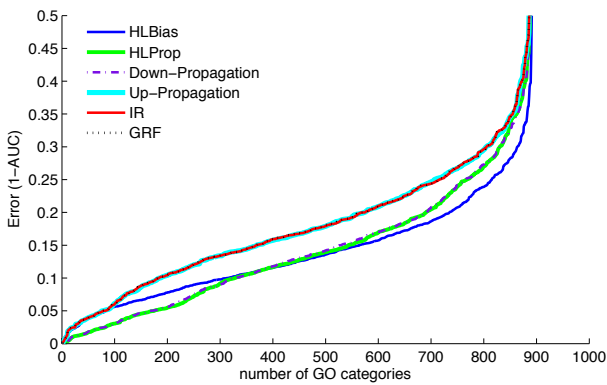

Figure 5: Comparison of performance in terms of error in predicting novel genes in 903 GO categories that acquired three or more annotations in a span of one year.

Figure 4 shows a typical set of ROC curves; these curves were generating for predicting GO category "Osteoblast Differentiation". Figure 5 shows the cumulative performance of each method in predicting novel genes in 903 categories as measured by error (1-AUC of ROC). Table 2 shows the mean and median error. As shown in Figure 5, unlike the cross-validation setting, the performance of HLProp at high percentiles (e.g. greater than 75% percentile) is better that that of HLBias. In particular, we found that our assumption about using genes that are annotated in siblings' categories as negative example may not always hold. For example, in 13 of the 903 GO categories, at least one newly annotated gene had a previous annotation in one of the sibling categories. The performance of HLBias (mean error of 0.1406) was worse than HLProp in these 13 categories (mean error of 0.1091) and similar to the baseline (mean error of 0.1469). In addition, some of the newly annotated genes had no previous annotations in ancestral categories with less than 300 annotations; as we discuss later, the performance of HLBias was degraded in predicting these categories.

Table 2: Mean and median error in predicting novel genes in 903 GO categories. The last column shows the standard error.

| Approach | mean | median | SE |
|---|---|---|---|
| Hierarchical label bias | 0.1367 | 0.1249 | 0.0025 |
| HLProp | 0.1404 | 0.1262 | 0.0032 |
| Down-Propagation | 0.1415 | 0.1305 | 0.0032 |
| Up-Propagation | 0.1767 | 0.1671 | 0.0031 |
| IR | 0.1768 | 0.1672 | 0.0031 |
| GRF | 0.1772 | 0.1673 | 0.0031 |

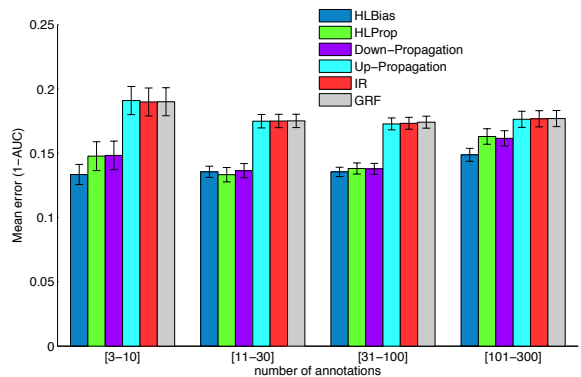

Figure 6: Comparison of performance in terms of error (1-AUC of ROC) in predicting novel genes. The performance is measured in predicting GO categories with [3-10] (136 categories), [11-30] (350 categories), [31-100] (314 categories), and [101-300] (101 categories) positive annotations in the 2008 GO file.

To better understand the difference between the various methods, we measured the mean performance in predicting GO categories at four different specificity levels; those with 3-10, 11-30, 31-100, and 101-300 annotations in the 2008 GO file. As shown in Figure 6, HLBias performs better than the others in predicting GO categories with 3 to 10 and 101 to 300 annotations whereas the performance of HLProp, Down-propagation, and GO label bias is similar when predicting GO categories with 11 to 100 annotations. Furthermore, using hierarchical information yields the most improvement on GO categories with a very few positive examples.

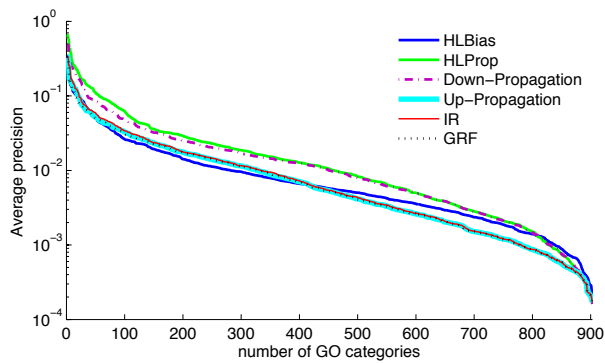

Figure 7: Comparison of performance in terms of average under the precision-recall curve when predicting novel genes in 903 GO categories that acquired three or more annotations in a span of one year.

In addition to measuring performance in terms of 1-AUC of ROC, we also investigated the performance in predicting novel gene functions in terms of area under the precision recall curve. As shown in Figure 7, the performance of HLProp in terms of average precision, is better than all other methods. Interestingly, in contrast to the ROC measure, we observed that on average, the performance of HLBias is not significantly different than the baseline approach (t-test with $\alpha=0.05$). However, the cumulative performance of HLBias follows the same trend as measured in error or average precision (compare Figure 5 and 7); HLBias has a lower precision at high percentile but higher precision at lower percentiles. The lower performance of HLBias, as measured in terms of precision, can be explained by our observation that genes with new annotations in 338 of the 903 novel categories had no previous annotation in the corresponding ancestral categories and were therefore deemed negatives and given a highly negative label bias: the performance of HLBias in terms of precision at lower recalls (e.g. recall of 10%) was specifically degraded in these categories.

## 6 Discussion

Here we have shown that by using the GO hierarchy information directly, either by setting initial label biases using GO or using our formulation of hierarchical Gaussian random fields (HLProp), we can significantly improve gene function prediction. On the other hand, our results are consistent with the previous report that reconciliation methods may rarely improve the performance of independent classifiers (Obozinski et al., 2008). In contrast, in our setting, reconciliation of independent GO category results in a performance very similar to the baseline of un-corrected classifications obtained by GRF.

In order to be able to solve HLProp efficiently, we ignored the directionality of the GO hierarchy. To do so, we set $p_{mc} = p_{cm}$ if category $c$ is a child of category $m$. In contrast, the two heuristics variants (Up- and Down-propagation) only propagate information about discriminant scores in one direction. Our results indicate that propagating information down the hierarchy results in most gain whereas Up-Propagation does not significantly effect the performance. This result is consistent with that of (Obozinski et al., 2008) which found that reconciliation with a Bayesian network model of GO hierarchy where the arrows are directed from parents to the children classes performs better than the opposite model where the arrows are directed from children to the parents.

In its most general form, the GRF and the HLProp algorithms contain a number of regularization parameters. We have had general success by setting all these parameters to one though we have not rigorously investigated the effect of changing them because performing cross-validation on genome scale datasets is computationally costly and sometimes infeasible. However, a future area of improvements may be to find an optimal setting for these parameters.

### Acknowledgements

We would like to thank David Warde-Farley for helpful comments. This work was supported by an NSERC operating grant, a Genome Canada grant administered by OGI and a CFI/MRI-LOF equipment grant to QM. SM was partially supported by an OGS fellowship.